\documentclass[10pt,twocolumn,letterpaper]{article}

\usepackage{cvpr}              
\usepackage{tabularx}
\usepackage{adjustbox}
\usepackage{makecell}

\usepackage[dvipsnames]{xcolor}

\definecolor{cvprblue}{rgb}{0.21,0.49,0.74}
\usepackage[pagebackref,breaklinks,colorlinks,citecolor=cvprblue]{hyperref}

\def\paperID{341} 
\def\confName{3DV\xspace}
\def\confYear{2024\xspace}

\title{Cross3DVG: Cross-Dataset 3D Visual Grounding on Different RGB-D Scans}

\author{
    \makecell{
        Taiki Miyanishi$^{1,3}$, \hspace{3pt}
        Daichi Azuma$^{2}$, \hspace{3pt}
        Shuhei Kurita$^{3}$, \hspace{3pt}
        Motoaki Kawanabe$^{1}$, \hspace{3pt}
        \\
        $^{1}$ATR, \hspace{3pt}
        $^{2}$Kyoto University, \hspace{3pt}
        $^{3}$RIKEN AIP
    }
}

\begin{document}
\maketitle
\begin{abstract}
We present a novel task for cross-dataset visual grounding in 3D scenes (Cross3DVG), which overcomes limitations of existing 3D visual grounding models, specifically their restricted 3D resources and consequent tendencies of overfitting a specific 3D dataset.
We created RIORefer, a large-scale 3D visual grounding dataset, to facilitate Cross3DVG. It includes more than 63k diverse descriptions of 3D objects within 1,380 indoor RGB-D scans from 3RScan~\cite{Wald2019RIO}, with human annotations.
After training the Cross3DVG model using the source 3D visual grounding dataset, we evaluate it without target labels using the target dataset with, e.g., different sensors, 3D reconstruction methods, and language annotators.
Comprehensive experiments are conducted using established visual grounding models and with CLIP-based multi-view 2D and 3D integration designed to bridge gaps among 3D datasets.
For Cross3DVG tasks, (i) cross-dataset 3D visual grounding exhibits significantly worse performance than learning and evaluation with a single dataset because of the 3D data and language variants across datasets. Moreover, (ii) better object detector and localization modules and fusing 3D data and multi-view CLIP-based image features can alleviate this lower performance.
Our Cross3DVG task can provide a benchmark for developing robust 3D visual grounding models to handle diverse 3D scenes while leveraging deep language understanding.\footnote{Project page: \url{https://github.com/ATR-DBI/Cross3DVG}}
\end{abstract}

%
%
\begin{figure}[hbtp]
 \centering
 \includegraphics[keepaspectratio, scale=0.5]
      {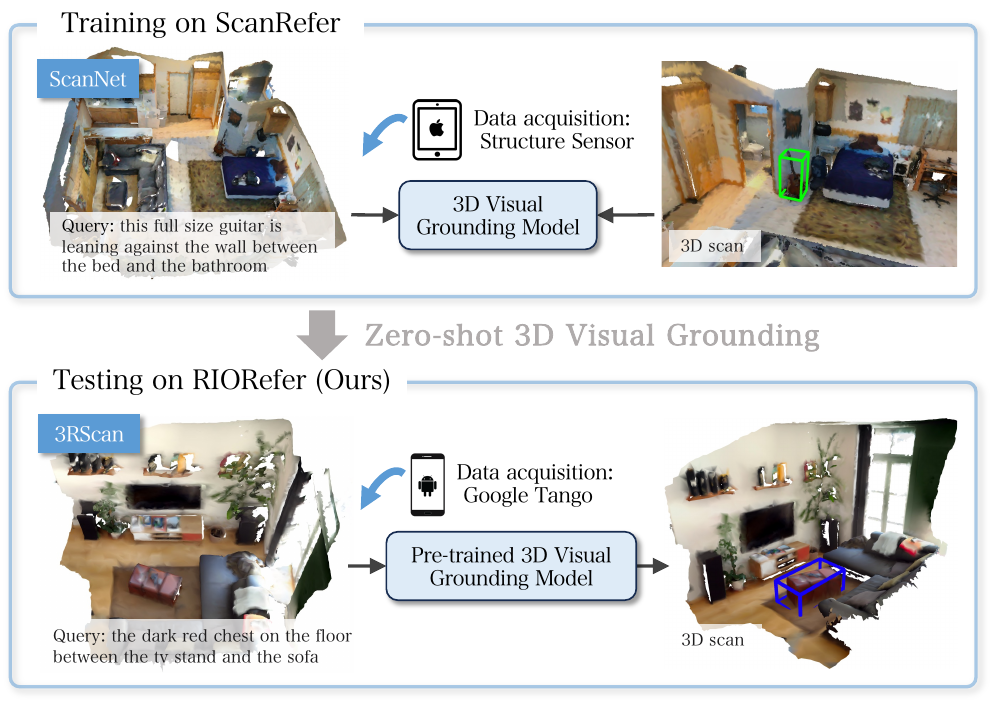}
\vspace{-0.4cm}
 \caption{
\small{
Example of cross-dataset 3D visual grounding with our Cross3DVG dataset consisting of ScanRefer and RIORefer.
We evaluate the cross-dataset 3D visual grounding model trained on one dataset using the other dataset.
}}
 \label{fig:teaser}
 \vspace{-0.4cm}
\end{figure}

\section{Introduction}
\label{sec:intro}
The rapid proliferation of affordable 3D scanners and remarkable advancements in 3D sensing methods have greatly accelerated the field of 3D object recognition~\cite{Chen_HAIS_2021_ICCV,jiang2020pointgroup,misra2021-3detr,Qi_2019_ICCV,rozenberszki2022language,Schult22_mask3d,vu2022softgroup},
with enhanced capabilities to assign semantic and instance labels to scanned point cloud data.
Leveraging 3D object recognition networks as backbones has facilitated the exploration of new research areas to connect languages and the 3D visual world, including 3D visual grounding~\cite{achlioptas2020referit_3d,chen2020scanrefer}, 3D dense captioning~\cite{chen2021scan2cap,ijcai2022p194,Yuan_2022_CVPR,Chen_2023_CVPR}, and 3D visual question answering~\cite{Azumascanqa,ma2022sqa3d}, thereby supporting elucidation of the real world through language.
Among these tasks, 3D language grounding stands as a pioneering effort in the realm of 3D vision and language, aimed at localizing objects in a 3D scene corresponding to given natural language descriptions.
In recent years, numerous methods have been proposed to tackle this challenging task~\cite{chen2022language,Huang_Lee_Chen_Liu_2021,huang2022multi,jain2022bottom,Yuan_2021_ICCV,Zhao_2021_ICCV,Wu_2023_CVPR}.
Nevertheless, despite its promising applications in AR/VR and personal robotics, the capabilities of current vision systems to perform 3D visual grounding remain inadequate because of the limited availability of linguistically annotated 3D resources.

To benchmark the 3D visual grounding performance, existing works rely primarily on two standard datasets, ScanRefer~\cite{chen2020scanrefer} and ReferIt3D~\cite{achlioptas2020referit_3d}, which assign a language description to each object in the ScanNet dataset~\cite{scannetdata}.
Despite the diversity of 3D data available with the advent of various 3D scanning systems (e.g., RealSense, Android's ARCore, and iPhone's ARKit), current 3D visual grounding methods are evaluated on a single indoor 3D dataset (i.e., ScanNet), which is acquired with a specific scanner and scanning system.
This limitation suggests that the 3D visual grounding method might be adversely affected by overfitting a single dataset, reflecting similar challenges to those encountered with 2D scenarios~\cite{subramanian-etal-2022-reclip,teney2016zero,Yang_2021_ICCV,Zhang_2021_CVPR}.
Consequently, cross-dataset generalization of 3D visual grounding is still not well-explored within 3D vision and language.
To address this new challenge, another large-scale 3D visual grounding dataset must be produced for use in combination with the existing one and with benchmarks of 3D visual grounding across both datasets.

This paper addresses the cross-dataset 3D visual grounding (Cross3DVG) task, which involves the performance of 3D visual grounding across 3D datasets acquired from different 3D scanning systems and language annotations. Figure~\ref{fig:teaser} presents an example of the task.
To this end, we create a new large-scale 3D visual grounding dataset that includes 63k different linguistic annotations assigned to 3D objects within 1,380 RGB-D indoor scans from the 3RScan dataset~\cite{Wald2019RIO}, which exceeds the sizes of ScanRefer~\cite{chen2020scanrefer}, and existing 52k linguistic annotations based on 800 indoor RGB-D scans of ScanNet.
In addition, to realize a 3D visual grounding method that is robust to changes in the 3D data, we propose a CLIP-based baseline model for the Cross3DVG task, which simultaneously uses multi-view 2D images and 3D point cloud data, incorporating CLIP prior information~\cite{radford2021learning}.
We assumed that robust 3D visual grounding can be achieved by initially detecting objects with 3D data and subsequently using the surrounding multi-view 2D images, which are less dependent on 3D data differences.

To evaluate generalized capabilities across the two 3D visual grounding datasets, we use standard and state-of-the-art visual grounding methods and our multi-view 2D and 3D-based baseline method.
After training these models on the source dataset, we subsequently tested them on the target dataset using no training data for the target.
The results demonstrate that (i) current 3D visual grounding methods perform significantly worse than learning and evaluation in a single dataset and show
(ii) marked room for improvement of performance remains in current methods compared to the use of Oracle object detection and object localization modules separately.
These findings suggest that 3D visual grounding requires more robustness for cross-dataset generalization.
Consequently, our cross-dataset 3D visual grounding task offers new possibilities for elucidating details of the 3D visual world through language.

Our contributions can be summarized as presented below.
\vspace{-0.2\baselineskip}      
{
\setlength{\leftmargini}{10pt}  

\begin{itemize}
	\setlength{\itemsep}{5pt}   
	\setlength{\parskip}{0pt}   
	\setlength{\itemindent}{0pt}
	\setlength{\labelsep}{5pt}  
\item We create a new 3D visual grounding dataset to explore a challenging cross-dataset 3D visual grounding benchmark using two indoor RGB-D scan datasets with large-scale language annotation, thereby revealing important limitations of current 3D visual grounding systems.
\item As a baseline method for the Cross3DVG task, we develop a 3D visual grounding model that connects 3D point clouds and multi-view 2D images around the detected 3D objects using CLIP.
\item We evaluate existing methods on two 3D visual grounding datasets extensively, thereby providing detailed comparisons and the difficulty of the Cross3DVG task.
The experimentally obtained results demonstrate the capabilities of improved detectors and object localization modules. The fusion of multi-view 2D and 3D data alleviates the gap separating different 3D visual grounding datasets.
\end{itemize}
}
\vspace{-0.2\baselineskip}

%
%
\begin{table}[t!]
\begin{center}
\scriptsize
\setlength{\tabcolsep}{4.7pt}
\begin{tabular}{lccc}

    \toprule
    Dataset   &  Acquisition / Environment & Data format & \#desc.\\
    \midrule
    REVERIE~\cite{qi2020reverie}    & Matterport cam. / Matterport3D~\cite{Matterport3D}   & image & 21,702 \\
    SUN-Spot~\cite{Mauceri_2019_ICCV}  & Kinect v2 / SUN RGB-D~\cite{Song_2015_CVPR}    & image & 7,990\\ 
    SUNRefer~\cite{liu2021refer}  & Kinect v2 / SUN RGB-D~\cite{Song_2015_CVPR}    & image & 38,495 \\
    Nr3D~\cite{achlioptas2020referit_3d} & Structure Sensor / ScanNet~\cite{scannetdata}     & 3D scan & 41,503 \\
    ScanRefer~\cite{chen2020scanrefer}  & Structure Sensor / ScanNet~\cite{scannetdata}     & 3D scan & 51,583 \\
    \textbf{RIORefer} (ours) & \textbf{Tango / 3RScan~\cite{Wald2019RIO}} & \textbf{3D scan} & \textbf{63,602} \\
    \bottomrule
\end{tabular}
	\vspace{-0.2cm}
    \caption{\small{Comparison of our proposed RIORefer dataset with existing visual grounding datasets of indoor data.}}
    \label{table:datasets}
\end{center}
\vspace{-0.5cm}
\end{table}

\section{Related Work} \label{sec:related_work}
\noindent
\textbf{Datasets for 3D Visual Grounding.}
To benchmark cross-dataset 3D visual grounding,
we create the RIORefer dataset, with annotated descriptions to indoor 3D scenes from 3RScan~\cite{Wald2019RIO}.
Table~\ref{table:datasets} presents a comparison of existing visual grounding datasets and our RIORefer dataset.
In general, 3D visual grounding datasets are created with language annotating existing 3D datasets, such as Matterport3D~\cite{Matterport3D}, SUN RGB-D~\cite{Song_2015_CVPR}, and ScanNet~\cite{scannetdata}.
These 3D datasets can be created using different 3D data acquisition methods such as Matterport camera, Kinect, and Structure Sensor.
REVERIE~\cite{Qi_2020_CVPR}, a large-scale remote object grounding dataset, includes 21k natural language descriptions for instructing embodied agents to locate target objects in photorealistic 3D indoor environments.
SUN-Spot~\cite{Mauceri_2019_ICCV} and SUNRefer~\cite{liu2021refer} specifically support object localization within single-view RGB-D images captured inside buildings or houses.
These datasets using RGB and RGB-D images from 3D environments. By contrast, Nr3D~\cite{achlioptas2020referit_3d} and ScanRefer~\cite{chen2020scanrefer} use dense 3D scans of ScanNet captured using Structure Sensor on an iPad.
However, despite real-world applications often employing various sensors, existing 3D visual grounding datasets mainly comprise 3D data acquired using a single type of 3D scanning system.

\begin{figure*}[htbp]
    \begin{subfigure}{0.34\linewidth}
        \centering
    \scriptsize
    \begin{tabular}{lcc}
    \toprule
     & \textbf{ScanRefer} & \textbf{RIORefer} \\
    \midrule
    \# descriptions & 51,583 & 63,602 \\
    \# scans        & 800  & 1,380 \\
    \# objects      & 11,046 & 31,801 \\
    \# objects per scan & {64.48} & {46.09} \\
    \# of descriptions per object & {4.67} & {2.0} \\
    Size of vocabulary & {4,197} & {6,952} \\
    Avg. len. of descriptions & {20.27} & {14.78} \\
    \bottomrule
	\end{tabular}
        \caption{Dataset statistics}
        \label{tab:table1}
    \end{subfigure}
    \begin{subfigure}{0.38\linewidth}
        \centering
        \includegraphics[keepaspectratio, scale=0.28]{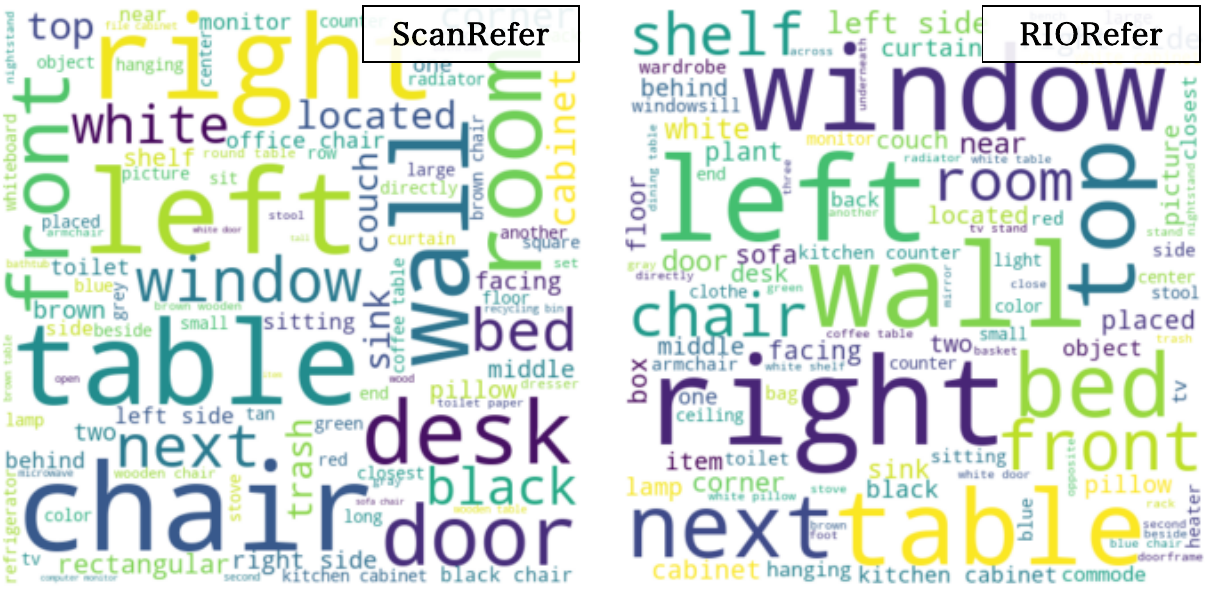}
        \caption{Word clouds}
        \label{fig:figure1}
    \end{subfigure}%
    \begin{subfigure}{0.25\linewidth}
        \centering
        \includegraphics[keepaspectratio, scale=0.29]{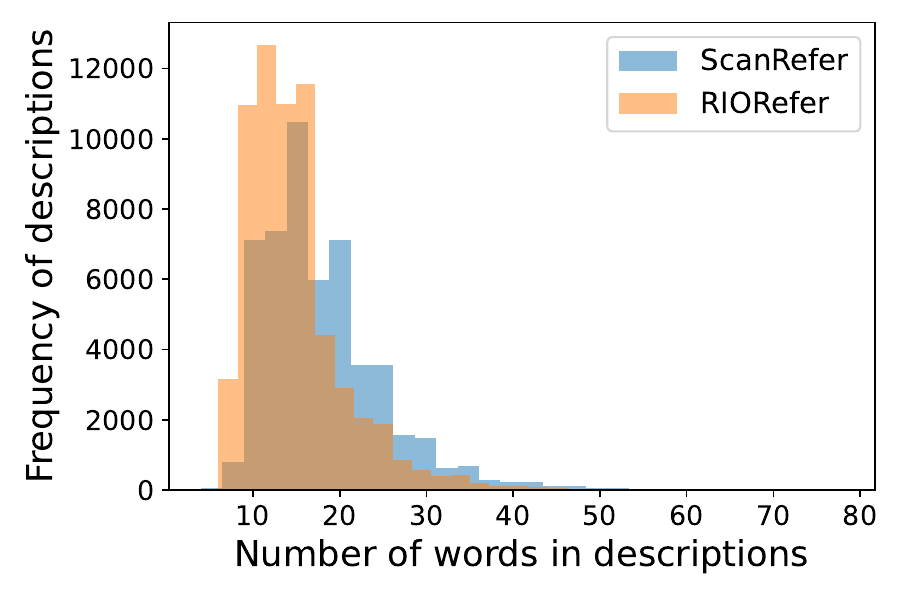}
        \caption{Description length}
        \label{fig:figure1}
    \end{subfigure}%
    \vspace{-0.2cm}
    \caption{Mutual differences of ScanRefer and RIORefer datasets.}
    \label{fig:diff_data_stats}
     \vspace{-0.5cm}
\end{figure*}

\vspace{0.1cm}
\noindent
\textbf{Learning Visual Grounding of 3D Scenes.}
The availability of standardized 3D visual grounding datasets has bolstered numerous insightful methods in this field.
Compared to 2D visual grounding~\cite{refcoco,refcocog} and video visual grounding~\cite{lingualotb99,VID-Sentence,Kurita_2023_ICCV} tasks for which state-of-the-art models~\cite{mdetr,ofa,liu2023grounding} use image encoders that are preliminary trained with large-scale image data, 3D visual grounding models use a limited number of 3D scene data.
Indeed, most existing 3D visual grounding methods use end-to-end learning of a 3D object detector and a localization module that combines language and 3D representations~\cite{bakr2022look,he2021transrefer3d,luo20223d,yuan2022toward,roh2022languagerefer,abdelreheem2022scanents,wang2023distilling}.
Some methods pre-compute 3D object detection or instance segmentation result. The methods then use the point cloud features of that 3D bounding box and segments for visual grounding~\cite{Huang_Lee_Chen_Liu_2021,Yuan_2021_ICCV}.
However, the low resolution of 3D data, resulting from the reconstruction process, presents challenges for object recognition.
To address this issue, using both 2D images and 3D data has been proposed~\cite{yang2021sat,jain2022bottom}.
Our baseline method differs from these methods because it can simultaneously perform object detection, acquire multi-view 2D images, and localize objects with multi-view CLIP-based 2D image and 3D object features.
Recently, some proposed methods can combine 3D visual captioning and grounding and learn both models concurrently ~\cite{cai20223djcg,chen2021d3net,chen2022unit3d,Jin_2023_CVPR}.
These works suggest that the simultaneous learning of both tasks exhibits synergistic effects.
To benchmark 3D visual grounding performance, most existing methods use two popular datasets based on ScanNet: Nr3D~\cite{achlioptas2020referit_3d} and ScanRefer~\cite{chen2020scanrefer}.
Hence the performance of 3D visual grounding across different 3D datasets remains unexplored.
For the work described herein, we employ established 3D visual grounding methods along with the CLIP-based baseline model (Section~\ref{sec:method}) to explore the components which contribute to generalization of the Cross3DVG task.

%
%
\begin{figure*}[hbtp]
 \centering
 \includegraphics[keepaspectratio, scale=0.72]
    {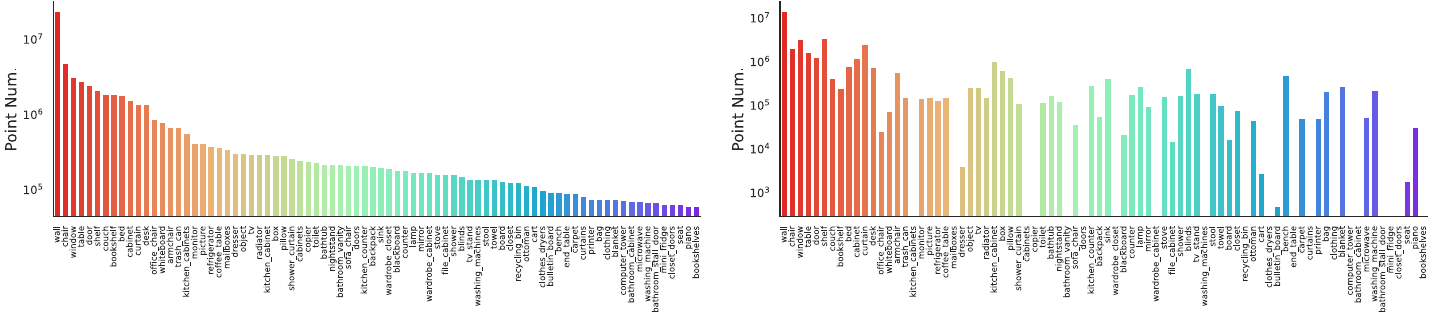}
\vspace{-0.5cm}      
 \caption{\small{Class category distributions for ScanRefer and RIORefer, based on the numbers of surface point annotations per category. Categories are ordered by the number of points per category on ScanRefer.}}
 \label{fig:point_num}
\vspace{-0.2cm}       
\end{figure*}

\section{Cross-dataset 3D Visual Grounding} \label{sec:dataset}
\subsection{Task Definition} \label{sec:task_definition}
We first provide an overview of the Cross3DVG task: given an input of the 3D scene and its description,
the model is designed to detect the 3D object which semantically matches the description.
For this work, we specifically assess the zero-shot setting across different 3D visual grounding datasets to elucidate the model's ability to generalize and adapt to novel target datasets, simulating real-world scenarios (e.g., robotics and augmented reality) in which the model encounters new 3D scenes captured by different 3D scanning systems.
For the zero-shot setting, using no training data for the target, we apply a pre-trained model learned on the source dataset to a target dataset.

\subsection{Dataset Construction} \label{sec:riorefer}
For the Cross3DVG task, we introduce the RIORefer dataset, which is comparable to ScanRefer.
Initially, RIORefer was created by annotating linguistic descriptions onto the 3RScan dataset~\cite{Wald2019RIO}, which includes 1,482 RGB-D scans of 478 indoor environments and which was created using Google Tango, which differs from the scanning system used for the creation of ScanNet.
Because one primary objectives of this study revolves around assessing the effects of different 3D data on the 3D visual grounding performance across datasets, we use ScanRefer~\cite{chen2020scanrefer} as a pair of RIORefer, rather than using Nr3D~\cite{achlioptas2020referit_3d}, which assumes the ground truth objects as given.

\vspace{0.1cm}
\noindent
\textbf{Annotation.}
To create the new 3D visual grounding dataset, we use a crowdsourcing service for manual annotation of descriptions to the axis-aligned 3RScan dataset.
First, we collect descriptions of objects across all 3D scenes using the interactive visualization website,
similarly to the dataset creation process described for ScanRefer (as described in Supplemental Materials of~\cite{chen2020scanrefer}).
We provide the following specific instructions to the worker for the object descriptions.
(i) Describe the object in the 3D scene so that the object can be uniquely identified in the scene based on the worker's description.
(ii) Include details about the object's appearance, location, and position relative to other objects in the description.
(iii) Record the viewpoint from which the object is viewed to describe the spatial relation among objects within the scene.

\vspace{0.1cm}
\noindent
\textbf{Verification and re-annotation.}
To improve the annotation quality, we filter out inappropriate descriptions using a manual 3D visual grounding website and re-annotate them after collecting descriptions.
After collecting the initial descriptions, we present the 3D scene and the corresponding object names and IDs to the workers.
The workers are subsequently instructed to enter the object IDs which best match the description provided for the scene. Furthermore, they are prompted to check a box if no object in the 3D scene matches the description, or if multiple objects correspond.
We discarded incomplete descriptions and re-annotated the corresponding objects using the annotation website used during the initial annotation step.
We then collected two descriptions for each object to capture multiple perspectives and linguistic variations.
We conducted a human performance assessment to evaluate the RIORefer dataset quality. Related details are given in the appendix.

\subsection{Dataset Analysis} \label{sec:data_stats}
Using crowdsourcing services, we collected 63,602 descriptions for approximately 249 unique objects across 1,380 scans as a RIORefer dataset.
To highlight differences between ScanRefer for RIORefer, we present some statistics, word clouds, and distributions of description lengths of the ScanRefer and RIORefer datasets in Fig.~\ref{fig:diff_data_stats}.
Although both distributions of description lengths and the terms used for the descriptions show similarities, the RIORefer includes more descriptions and a widely diverse vocabulary.
Furthermore, the numbers of annotated surface points distribution per category are provided for both datasets in Figure~\ref{fig:point_num}.
ScanRefer and RIORefer have different point distributions for each object.
Moreover, certain objects which frequently appear in ScanRefer do not appear in RIORefer.
These findings underscore the difficulty of cross-dataset 3D visual grounding: it requires grounding on objects that are scarce or which exhibit different point cloud sizes in the training data.

%
%
\begin{figure*}[hbtp]
\vspace{0.2cm}
 \centering
 \includegraphics[keepaspectratio, scale=0.76]
      {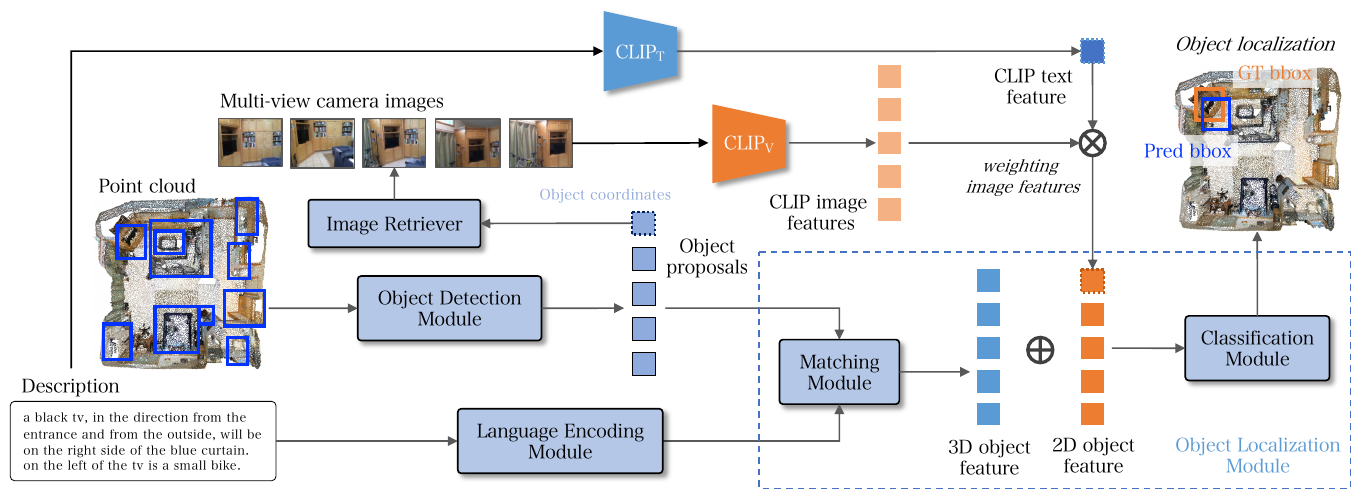}
\caption{\small{Overview of the proposed method.
 Our method takes as input a 3D point cloud and RGB images of a scene.
For each object proposal predicted by the object detector module, the method retrieves the surrounding object images using the camera pose matrix and the bounding box coordinates. 
The method assigns weights to image features extracted from the surrounding object images by the CLIP image encoder (CLIP$_\mathrm{I}$) using text features encoded from the description by the CLIP text encoder (CLIP$_\mathrm{T}$).
Finally, the object localization module computes the object confidence scores using 2D and 3D object features, subsequently predicting the bounding box of the target object.
}}
 \label{fig:network_overview}
 \vspace{-0.2cm}
\end{figure*}

\section{Methods} \label{sec:method}
To explore the Cross3DVG task difficulty, we prepared several baseline models based on different components.
Baselines include an established 3D visual grounding model ScanRefer~\cite{chen2020scanrefer} and 3DVG-Transformer~\cite{Zhao_2021_ICCV}\footnote{For our experiments, we used a follow-up work 3DVG-Transformer+ for our experiment, which incorporates a different spatial proximity matrix in the localization module. This module is used as a listener model in an earlier study~\cite{chen2021d3net}.}.
To construct a robust baseline model,
we extended the 3DVG-Transformer by incorporating CLIP-based multi-view features~\cite{radford2021learning}.
This extension enables the fusion of features extracted from both the multi-view 2D image and 3D data, specifically addressing differences between the two 3D datasets and providing a fair comparison of components with those of the existing method.

\vspace{0.1cm}
\noindent
\textbf{Overview of Network Architecture.}
Figure~\ref{fig:network_overview} portrays the overall architecture of our proposed baseline model, which comprises several key modules including language encoding, object detection, localization, and a classifier.
After the model takes the point cloud representation of the 3D scene as input, it uses a 3D object detector to predict object proposals within the scene.
Next, after the model takes a linguistic description as input, it computes how well the object proposals align with the linguistic cues for a localization module.
Unlike traditional transformer-based 3D visual grounding models, our approach enriches the grounding process by incorporating both multi-view 2D image information and 3D geometric information.
Finally, a classifier module computes the localization scores for the proposed object boxes, using the fusion of both the multi-view 2D and 3D object features.

\vspace{0.1cm}
\noindent
\textbf{Language Encoding Module.}
To ensure a fair comparison, we adopt a language encoding method similar to that used for ScanRefer.
First, we encode the words in the description using GloVe~\cite{pennington2014glove}, obtain word representations, and feed them into a one-layer GRU~\cite{Chung-et-al-TR2014} for word sequence modeling.
We obtain the contextualized word representation, denoted as $ w \in R^{d \times n}$, where $n$ represents the number of words in the description and $d$ stands for the hidden size of the GRU (set as 128).
The description is also converted to a sentence representation, denoted as $t \in R^{c}$ using CLIP's text encoder~\cite{radford2021learning} for weighting image features as described in the subsequent steps, where $c$ represents the dimensions of the CLIP model (set as 728).

\vspace{0.1cm}
\noindent
\textbf{Object Detection Module.}
A transformer-based object detector DETR3D~\cite{zhao2021transformer3d} is used. It serves as a backbone module of 3DVG-Transformer to extract point cloud feature ${p} \in R^{d}$
with a PointNet++~\cite{qi_2017_NIPS}. Then we apply a voting method for feature aggregation that considers multi-level relations among neighboring objects (xyz coordinates and normals are used for each point.)
The detector outputs object proposals (object bounding boxes) and their object features $f \in R^{d \times m}$,
where $m$ denotes the number of object proposals (set as 256.)
It is noteworthy that an alternative object detector such as VoteNet~\cite{Qi_2019_ICCV} can be used instead of DETR3D.

\vspace{0.1cm}
\noindent
\textbf{Matching Module.}
Following the architecture of 3DVG-Transformer, we use a transformer~\cite{NIPS2017_3f5ee243} to associate 3D and language features using an attention mechanism.
This approach enables the creation of a language-aware-object proposal representation $o_{3d} \in R^{d \times m}$ by feeding the language feature and object proposal feature into transformer layers.
In contrast to the vanilla 3DVG-Transformer, our baseline model uses both multi-view 2D and 3D object features.
Actually, 3D visual grounding methods relying solely on 3D data might be vulnerable to dataset shift
attributable to variations in the numbers of points in objects across different 3D datasets, as discussed in Section~\ref{sec:data_stats}.
We assumed (i) multi-view 2D images around a bounding box that include a rich object context and (ii) 2D image features extracted using a model trained on numerous images that are independent of 3D data
and which exhibit robustness to 3D dataset shifts.

After extracting image features $v \in R^{c}$ from the RGB camera images used to create the 3D scan with the CLIP image encoder to incorporate these assumptions, we identify 2D images related to the 3D object. Then we compute the cosine similarity between the center coordinates of the predicted object proposals and the camera matrix of each image.
We select the top $l$ (set as 10) multi-view images close to the object proposal with the highest cosine similarity values, provided that their cosine similarity exceeds a specified threshold (empirically set as 0.8).
Finally, we compute the weighted image features by the inner product value of text feature $t$ extracted by the CLIP text encoder, and by reducing the weight of irrelevant images based on the given description.
By projecting weighted average image features, we obtain the 2D object features $o_{2d} \in R^{d\times m}$.
In contrast to the existing methods using 2D image and 3D object features,
which rely on a single image of the proposed object ~\cite{yang2021sat,jain2022bottom},
our method can use multi-view 2D image features for learning end-to-end 3D visual grounding without requiring a ground truth image or label.

\vspace{0.1cm}
\noindent
\textbf{Classifier Module.}
To assess the relevance of an object to the given description, we compute a localization score with ${s}_{j} = \mathrm{Softmax}(\mathrm{GELU}(\mathrm{MLP}([{o}_j])))$.
Here, $\mathrm{GELU}$ refers to the Gaussian error linear unit activation function~\cite{Chung-et-al-TR2014},
$o_j$ represents the object features, and ${s}_j \in \mathbb{R}^{1 \times m}$ stands for the softmax scores of object proposals.
Specifically, we compute the localization scores $s_{2d}$, $s_{3d}$, and $s_{2d3d}$ using the 2D object feature $o_{2d}$, language-aware 3D object feature $o_{3d}$ and their joint object feature $o_{2d} + o_{3d}$.
Finally, we select the target object with the highest $s_{2d3d}$ scores.

\vspace{0.1cm}
\noindent
\textbf{Loss Function.}
To train our model, we use a loss function resembling those used in ScanRefer and 3DVG-Transformer.
This loss function comprises the object classification loss $\mathcal{L}_{cls}$, object detection loss $\mathcal{L}_{det}$, and localization loss $\mathcal{L}_{loc}$,
where we use a combination of $\mathcal{L}_{loc}(o_{2d})$ + $\mathcal{L}_{loc}(o_{3d})$ + $\mathcal{L}_{loc}(o_{2d} + o_{3d})$ as $\mathcal{L}_{loc}$ to facilitate joint learning.
In line with the approach used for an earlier study~\cite{chen2021d3net}, we adopt a linear combination loss $\mathcal{L} = 0.1 \mathcal{L}_{cls} + 0.3 \mathcal{L}_{loc} + 10 \mathcal{L}_{det}$ as a final loss.

\subsection{3D Visual Grounding Baselines}
The following 3D visual grounding baselines are implemented.
Our primary emphasis is constructing a dataset for the Cross3DVG task and elucidating the components contributing to its success.
Therefore, complex models are not used, and not pure 3D visual grounding, such as~\cite{cai20223djcg} and~\cite{chen2021d3net}, which simultaneously perform 3D dense captioning.

%
%
\setlength{\tabcolsep}{2pt}
\begin{table*}[t]
\begin{center}
\small
	\begin{tabular}{lcccccc|cccccc}
        \toprule
    & \multicolumn{6}{c}{\scriptsize ScanRefer$\rightarrow$RIORefer} & \multicolumn{6}{c}{\scriptsize RIORefer$\rightarrow$ScanRefer} \\
    \midrule
    & \multicolumn{2}{c}{unique}  & \multicolumn{2}{c}{multiple}  & \multicolumn{2}{c}{overall} & \multicolumn{2}{c}{unique}  & \multicolumn{2}{c}{multiple}  & \multicolumn{2}{c}{overall} \\
Model  &{\scriptsize Acc@.25}  & {\scriptsize Acc@.5} & {\scriptsize Acc@.25}  & {\scriptsize Acc@.5} & {\scriptsize Acc@.25}  & {\scriptsize Acc@.5} & {\scriptsize Acc@.25}  & {\scriptsize Acc@.5} & {\scriptsize Acc@.25}  & {\scriptsize Acc@.5} & {\scriptsize Acc@.25}  & {\scriptsize Acc@.5} \\
\midrule
VoteNet+MLP  & 31.46 & 15.56 & 13.84 & 7.10 & 19.69 \textcolor[gray]{0.5}{(30.60)} & 9.91 \textcolor[gray]{0.5}{(16.54)} & 40.23 & 22.41 & 19.18 & 11.98 & 27.27 \textcolor[gray]{0.5}{(40.38)}& 15.99 \textcolor[gray]{0.5}{(27.04)}\\
VoteNet+Trans  & 32.98 & 17.10 & 16.89 & 9.34 & 22.23 \textcolor[gray]{0.5}{(33.52)} & 11.92 \textcolor[gray]{0.5}{(18.45)}& 45.06& 25.30 & 22.09 & 14.27& 30.92 \textcolor[gray]{0.5}{(45.08)} & 18.51 \textcolor[gray]{0.5}{(30.42)}\\
DETR3D+Trans  & 36.05 & 18.38 & 16.91  & 9.29 & 23.27 \textcolor[gray]{0.5}{(34.87)}& 12.31 \textcolor[gray]{0.5}{(19.41)} & 45.05 & 25.41 & 22.76 & 14.36 & 31.33 \textcolor[gray]{0.5}{(46.64)}& 18.60 \textcolor[gray]{0.5}{(31.83)}\\
Ours
& \textbf{37.37} & \textbf{19.77} & \textbf{18.52} & \textbf{10.15} & \textbf{24.78} \textcolor[gray]{0.5}{(36.56)} & \textbf{13.34} \textcolor[gray]{0.5}{(20.21)} & \textbf{47.90} & \textbf{26.03} & \textbf{24.25} & \textbf{15.40} & \textbf{33.34} \textcolor[gray]{0.5}{(48.74)} & \textbf{19.48} \textcolor[gray]{0.5}{(32.69)} \\
\bottomrule
	\end{tabular}
	\vspace{-0.2cm}
    \caption{
    \small{
Performance comparison of cross-dataset 3D visual grounding using baseline methods.
The `unique' category refers to cases for which only a single object of its class exists in the 3D scene. In contrast, the `multiple' category refers to cases for which multiple objects of its class exist in the 3D scene.
The gray numbers in the evaluation represent the performance values obtained when the training set and validation set are from the same dataset.
}}
    \label{table:result_s2r_r2s}
\end{center}
\vspace{-0.5cm}
\end{table*}

\noindent \textit{\textbf{Random}}:
To explore the difficulty of the Cross3DVG task, we prepare a random grounding method. It selects a target object's bounding box randomly from the object proposals.
Comparing the performance of this method with the other baselines yields some insights into the feasibility of zero-shot cross-dataset 3D visual grounding.

\noindent \textit{\textbf{OracleObjLoc}}:
This method selects the box from the object proposals predicted by the 3D object detector module, which best matches the ground truth.
Using this method, the Cross3DVG performance when object localization is correct can be ascertained. Then, it is possible to
evaluate the effect of textual differences alone in two datasets on 3D visual grounding compared to our baseline method.

\noindent \textit{\textbf{OracleObjDet}}:
This method uses ground truth object bounding boxes and selects the target by the trained localization modules.
Using this method, the Cross3DVG performance when 3D object detection is correct is apparent. It is possible to evaluate the effects of 3D data differences alone in the two datasets on 3D visual grounding.

\noindent \textit{\textbf{VoteNet+MLP (ScanRefer)}}~\cite{chen2020scanrefer}:
The ScanRefer is a well-established 3D visual grounding method that uses VoteNet~\cite{Qi_2019_ICCV} to take a point cloud as input and to predict object proposals within the scene. Subsequently, it fuses them with the encoded language feature of the input description and outputs the object localization scores for the object proposals using the multilayer perceptron (MLP).

\noindent \textit{\textbf{VoteNet+Trans}}:
To explore the effects of different localization modules on Cross3DVG,
we introduce a model using a transformer-based object localization module.
This introduced model enables us to compare its performance with the former
\textit{VoteNet+MLP} model and enables us to analyze how the choice of 3D object localization module influences Cross3DVG.

\noindent \textit{\textbf{DETR3D+Trans (3DVG-Transformer)}}~\cite{Zhao_2021_ICCV}:
This baseline method uses a transformer-based object detector DETR3D~\cite{zhao2021transformer3d} as its backbone and uses the transformer-based object localization module.
To ensure a fair comparison, we use the publicly available code of this method\footnote{\url{https://github.com/zlccccc/3DVG-Transformer}}.
From comparison of this method with our method, it is possible to use multi-view images to validate the CLIP module effectiveness.

\section{Results} \label{sec:experiment}
\subsection{Experiment setup}
\textbf{Implementation details.}
The publicly available codebases of ScanRefer~\cite{chen2020scanrefer} and 3DVG-Transformer+~\cite{Zhao_2021_ICCV} are used to implement our baseline models.
Similarly to earlier work~\cite{Zhao_2021_ICCV}, we train all models using AdamW optimizer~\cite{loshchilov2017decoupled}.
We mainly used the default parameter for learning rates on ScanRefer and 3DVG-Transformer.
Our baseline method used cosine annealing with a weight decay factor of 1e-5.
We set 200 epochs for training with a batch size of 8 and a chunk size of 32, which corresponds to 32 descriptions with eight scenes.
Additional training details are presented in an earlier report of the relevant literature \cite{Zhao_2021_ICCV}.

%
%
\setlength{\tabcolsep}{2pt}
\begin{table}[t]
\begin{center}
\small
	\begin{tabular}{lcccccc|cccccc}
        \toprule
    & \multicolumn{2}{c}{\scriptsize ScanRefer$\rightarrow$RIORefer} & \multicolumn{2}{c}{\scriptsize RIORefer$\rightarrow$ScanRefer} \\
    \midrule
Model & {\scriptsize Acc@.25}  & {\scriptsize Acc@.5} & {\scriptsize Acc@.25}  & {\scriptsize Acc@.5} \\
\midrule
OracleObjLoc &  \textcolor[gray]{0.5}{29.87} &  \textcolor[gray]{0.5}{28.71} &  \textcolor[gray]{0.5}{37.45} &  \textcolor[gray]{0.5}{36.79} \\
OracleObjDet &  \textcolor[gray]{0.5}{56.95} &  \textcolor[gray]{0.5}{30.22} &  \textcolor[gray]{0.5}{78.49} &  \textcolor[gray]{0.5}{53.81} \\
\midrule
Random   & 15.55 & 5.88 & 22.90 & 10.70\\
Ours & \textbf{24.78} & \textbf{13.34} & \textbf{33.34} & \textbf{19.48}\\
\bottomrule
	\end{tabular}
	\vspace{-0.2cm}
    \caption{
    \small{
Comparison of baseline methods to confirm the possibility of 3D visual grounding.
}}
    \label{table:task_difficulty_results}
\end{center}
\vspace{-0.6cm}
\end{table}

\vspace{0.1cm}
\noindent \textbf{Evaluation measures.}
To evaluate cross-dataset 3D visual grounding methods, we used the same evaluation metrics used in earlier ScanRefer work~\cite{chen2020scanrefer}: the percentage of predictions for which the Intersection over Union (IoU) with the ground truth boxes exceeds a certain threshold. Specifically, two thresholds were considered: IoU $>$ 0.25 and IoU $>$ 0.5.

\subsection{Cross-Dataset Visual Grounding Performance}
As described in this section, we evaluated the cross-dataset 3D visual grounding task, which involves training models on one dataset and evaluated their performance on another without access to manual visual and textual data labels during training.

\vspace{0.1cm}
\noindent
\textbf{Difficulty of cross-dataset 3D visual grounding.}
To elucidate the degree of challenge involved with cross-dataset 3D visual grounding, we present performance indicators of the proposed method with \textit{Random}, \textit{OracleObjDet}, and \textit{OracleObjLoc} in Table~\ref{table:task_difficulty_results}.
Our method performed better than the \textit{Random} baseline, randomly selecting a target box from the object detection results indicating that zero-shot cross-dataset 3D visual grounding is possible.
However, much room for improvement remains, as clarified by the results of the
\textit{OracleObjLoc} and \textit{OracleObjDet}.
\textit{OracleObjLoc} achieved notably superior results compared to our method by selecting an oracle object from the candidates.
Furthermore, \textit{OracleObjDet} exhibited remarkable superiority over our baseline method, indicating that the accuracy of 3D object detection across different datasets strongly affects the Cross3DVG task.
Although these results are based on ideal methods, the difficulty of Cross3DVG is also confirmed when employing more realistic methods, as presented in Table~\ref{table:result_s2r_r2s}.
The results show a marked decrease in accuracy for all methods when the datasets differed between the training and validation sets.
For example, the performance of \textit{Ours} drops from 36.56\% to 24.78\% for Acc@.25 for the RIORefer dataset and from 48.74\% to 33.34\% for the ScanRefer dataset.
These findings suggest the difficulty of cross-dataset 3D visual grounding and suggest the importance of developing robust 3D object detection and object localization modules to handle both 3D and language variations during training and testing.

%
%
\setlength{\tabcolsep}{2pt}
\begin{table}[t]
\begin{center}
\small
\begin{tabular}{lcc|cc}
\toprule
& \multicolumn{2}{c}{\scriptsize ScanRefer$\rightarrow$RIORefer} & \multicolumn{2}{c}{\scriptsize RIORefer$\rightarrow$ScanRefer} \\
\midrule
Model & {\scriptsize Acc@.25}  & {\scriptsize Acc@.5} & {\scriptsize Acc@.25}  & {\scriptsize Acc@.5} \\
\midrule
VoteNet+MLP (xyz only) & 18.97 & 9.51 &  25.80 & 14.81 \\
VoteNet+MLP w/ color & 18.79 & 9.64  & 25.50 & 14.70 \\
VoteNet+MLP w/ normal & 19.69 & 9.91 & 27.27 & 15.99 \\
\midrule
DETR3D+Trans (xyz only) & 22.41 & 11.86 &  30.17 & 17.99 \\
DETR3D+Trans w/ color & 22.22 & 11.58 & 29.35 & 17.09 \\
DETR3D+Trans w/ normal & 23.27 & 12.31   & 31.33 & 18.60 \\
\midrule
Ours (xyz only) & 24.01 & 12.70 & 31.74 & 18.77 \\
Ours w/ color  & 22.12 & 11.27 & 31.40 & 18.50 \\
Ours w/ normal & \textbf{24.78} & \textbf{13.34} & \textbf{33.34} & \textbf{19.48} \\
\bottomrule
\end{tabular}
\vspace{-0.2cm}
\caption{\small{Ablation study of the proposed baseline method with different features.}}
\label{table:result_feature}
\end{center}
\vspace{-0.4cm}
\end{table}

\vspace{0.1cm}
\noindent
\textbf{Effects of the CLIP-based multi-view image module.}
The \textit{Ours} method using the CLIP-based multi-view image module
trained on ScanRefer achieves the respective accuracies of 24.78\% and 13.34\% on RIORefer for the Acc@.25 and Acc@.5 measures.
Conversely, when trained on RIORefer, the method achieves accuracies of 33.34\% and 19.48\% on ScanRefer.
Furthermore, our method consistently outperformed \textit{DETR3D+Trans} which uses no CLIP module, across overall and even unique and multiple categories.
These findings highlight the potential of incorporating CLIP-based multi-view image features to improve the performance of 3D visual grounding models for accommodating dataset variations.

%
%
\setlength{\tabcolsep}{2pt}
\begin{table}[t]
\begin{center}
\small
	\begin{tabular}{lcc|cc}
        \toprule
    & \multicolumn{2}{c}{\scriptsize ScanRefer$\rightarrow$RIORefer} & \multicolumn{2}{c}{\scriptsize RIORefer$\rightarrow$ScanRefer} \\
    \midrule
Model  & {\scriptsize Acc@.25}  & {\scriptsize Acc@.5} & {\scriptsize Acc@.25}  & {\scriptsize Acc@.5} \\
\midrule
Ours wo/ text\_filter & 23.26 & 12.14 & 31.01 & 18.50 \\
Ours w/ obj\_name  & 22.74 & 11.81 & 33.51 & 19.84 \\
\midrule
Ours (frame 1)  & 23.47 & 12.35 & 32.52 & 19.34 \\
Ours (frame 5)  & 24.38 & 12.71 & 32.75 & 19.31 \\
Ours (frame 15) & 23.63 & 12.58 & 33.18 & {19.92} \\
\midrule
Ours & {24.78} & {13.34} & {33.34} & {19.48} \\
\bottomrule
	\end{tabular}
	\vspace{-0.2cm}
    \caption{\small{Ablation study of the proposed baseline method with different components.}}
    \label{table:result_ablation}
\end{center}
\vspace{-0.4cm}
\end{table}

\vspace{0.1cm}
\noindent
\textbf{Effects of better object localization modules.}
A better object localization module can illustrate the improvement of the cross-dataset 3D visual grounding.
We observed that \textit{VoteNet+Trans} consistently performed better than \textit{VoteNet+MLP} for both datasets.
This finding suggests that the transformer-based localization module, with integrated language and 3D features, is more effective for the Cross3DVG task than the MLP-based module.
Developing a better object localization module is expected to be crucially important for the success of Cross3DVG.

\vspace{0.1 cm}
\noindent
\textbf{Effects of better object detector modules}
Earlier research demonstrated the superiority of DETR3D over VoteNet for single dataset visual grounding~\cite{Zhao_2021_ICCV}.
However, the performance of these methods for cross-dataset 3D visual grounding datasets remains unexplored.
Motivated to answer this remaining question of the relevant literature, we compared \textit{VoteNet+Trans} and \textit{DETR3D+Trans}, which revealed that \textit{DETR3D+Trans} slightly outperforms \textit{VoteNet+Trans} on both datasets, suggesting that a better object detector can be effective for the Cross3DVG task to some degree.
However, this performance improvement was limited, in contrast to \textit{OracleObjDet}.
Based on these findings, our proposed dataset and the Cross3DVG task have revealed the challenge of this task, deriving specifically from differences in 3D data between ScanNet and 3RScan.

%
%
\begin{figure*}[hbtp]
\vspace{0.4cm}
 \centering
 \includegraphics[keepaspectratio, scale=0.7]
      {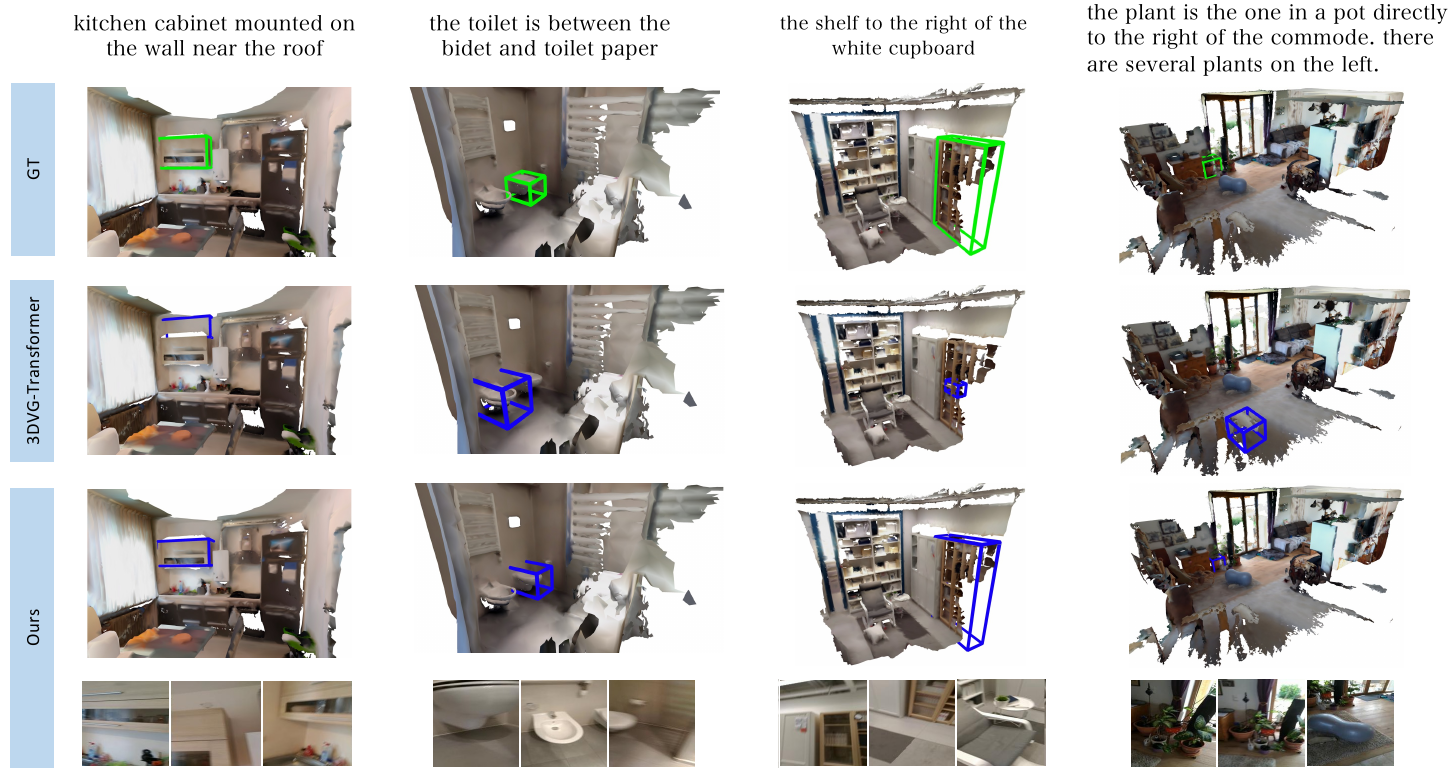}
\vspace{-0.1cm}
  \caption{\small{Qualitative results from 3DVG-Transformer and our method (Ours). The ground truth (GT) boxes are shows as green, whereas the predicted boxes with an IoU score higher than 0.5 are shown as blue.
 The text at the top of the image presents a given description for visual grounding.
 The additional images at the bottom of our method (Ours) are retrieved images based on the coordinates of object proposals.}}
\label{fig:qalitative}
\vspace{-0.3cm}
\end{figure*}

\subsection{Ablation study}
\noindent
\textbf{Effects of colors and normals.}
Many 3D visual grounding works used RGB (color) and normals of point clouds and showed better performance than that obtained when using only xyz coordinates~\cite{chen2020scanrefer,Yuan_2021_ICCV,Zhao_2021_ICCV}, but the effects of these features for the Cross3DVG task remain unexplored.
To investigate this point, we compared \textit{VoteNet+MLP}, \textit{DETR3D+Trans}, and \textit{Ours}
trained with geometry (xyz only) with a model trained with RGB values (w/ color) and a model trained with normal values (w/ normal).
The results presented in Table~\ref{table:result_feature} indicate that RGB values did not bring improvements across all models.
By contrast, normal consistently improved the performance of the models.
This observation suggests that the geometric information is robust and informative for the Cross3DVG task.
In contrast, color information (RGB values) is less reliable because color is more diverse across the different datasets, ScanNet and 3RScan.
To address this color variation, Cross3DVG models should incorporate robust image features extracted from models trained on web-scale image datasets, such as CLIP.

\vspace{0.1cm}
\noindent
\textbf{Effects of multi-view images.}
We examined the effects of using different numbers of images around the object on the performance of our CLIP-based method.
Multi-view images enabled us to capture context around objects and improve the overall view of large objects (e.g., a bed).
\textit{Ours} (frame $l$) in Table~\ref{table:result_ablation} presents the 3D visual grounding results obtained using different numbers of images.
The results demonstrate that using a single image, as in existing methods~\cite{yang2021sat,jain2022bottom}, does not yield satisfactory performance because it might lack sufficient context and viewpoint diversity.
In contrast, the performance improves as the number of images increases, capturing a richer contextual understanding of the objects and obtaining a more comprehensive view.
These findings underscore the importance of considering multiple viewpoints and incorporating contextual information from multiple images to improve the Cross3DVG performance.

\vspace{0.1cm}
\noindent
\textbf{Effects of image weighting with text.}
After investigating the effects of filtering image features based on textual features extracted from the description using the CLIP text encoder, we compared the proposed baseline model performance achieved with and without text filtering.
The results \textit{Ours} w/o text\_filter shown in Table~\ref{table:result_ablation} indicate
the importance of using textual information to guide the selection of relevant visual content.
Furthermore, we compared the effectiveness of using the object description vs. the object name for filtering.
The model employing a description-based image weighting approach performs better than using an object name (\textit{Ours} w/ obj\_name) on RIORefer. However, this approach leads to challenges on ScanRefer.
RIORefer's descriptions are shorter than those of ScanRefer, indicating that the CLIP text encoder can extract beneficial text features for filtering and improving performance effectively.

\subsection{Qualitative analysis}\label{sec:qualitative}
Figure~\ref{fig:qalitative} presents qualitative results obtained using 3DVG-Transformer and our proposed baseline method.
The results demonstrate that our method correctly predicts the target object according to the descriptions, but that 3DVG-Transformer fails to make that prediction.
Our method demonstrates a discriminative ability in cross-dataset visual grounding using the CLIP features.
For example, in the first column, 3DVG-Transformer incorrectly predicted the object proposal near the correct object (the `wall' above the kitchen cabinet).
However, our proposed method was able to identify the target object (`kitchen cabinet') using 2D images surrounding the target object.

\section{Conclusion} \label{sec:conclusion}
As described herein, we have introduced the Cross3DVG dataset, which serves as a benchmark for evaluating the performance of 3D visual grounding models trained on one dataset and tested on another. Moreover, we have provided detailed insights into the dataset creation process using crowdsourcing, as well as baseline experiments to elucidate models' behaviors and performance.
This benchmark dataset can contribute importantly to the advancement of 3D visual grounding research, particularly in the context of robust 3D understanding for real-world applications such as home robots and AR/VR systems.
These applications often operate in diverse environments and rely on various sensors, requiring models that are able to generalize well across different datasets and which can interpret spatial and semantic information through language.
This benchmark is expected to serve as a valuable resource for developing more robust and effective visual grounding models, thereby advancing the field of 3D understanding and its practical applications.

\section{Acknowledgments}
This work was supported by JST PRESTO JPMJPR22P8, and JPMJPR20C2, and by JSPS KAKENHI 22K12159 and 22K17983.

{
    \small
    \bibliographystyle{ieeenat_fullname}
    \bibliography{main}
}
\clearpage
\setcounter{page}{1}
\maketitlesupplementary

\section*{Appendix}
As presented in this appendix, we first present additional dataset details in Section~\ref{sec:supp_dataset}.
Then we provide experiment details in Section~\ref{sec:supp_exp_setup}.
We describe an additional ablation study in Section ~\ref{sec:supp_ablation} and additional qualitative results in Section~\ref{sec:supp_qualitative}.

\section{Additional Dataset Details}~\label{sec:supp_dataset}
To create the Cross3DVG dataset, we constructed the RIORefer dataset, which consists of more than 63k unique descriptions of 3D objects in 1,380 indoor RGB-D scans obtained from the 3RScan dataset. This section describes the pre-processing steps applied to the 3RScan data, the annotation website used for data collection, and an evaluation of human performance in 3D visual grounding.

\subsection{Dataset Construction}
%
%
\vspace{0.1cm}
\noindent
\textbf{Preprocess.}
We used an automatic axis alignment process to address the misalignment issue between the ground truth bounding boxes and the predicted ones in the 3RScan dataset. This process helps to ensure consistency and accuracy in the alignment of the 3D data. The axis-alignment procedure is the following:

\begin{enumerate}
\item We first identify the longest `wall' in the scene. If there are no walls, it looks for long objects that are typically parallel to the room such as "curtains," "windows," and "doorframes."

\item Once the longest wall or a suitable long object is identified, the 3D data are rotated along the wall's axis. This alignment step helps to align the data with the main structure of the scene.

\item Then, we adjust the 3D data height so that the horizontal and depth axes are grounded to the floor. This step ensures the scene's horizontal and depth information is aligned correctly with the ground plane.

\item The final step rotates the 3D data so that the vertical axis faces upward. This rotation ensures that the orientation of the objects is consistent and facilitates easier interpretation and analysis of the scene.
\end{enumerate}
By applying this automatic axis-alignment process, we mitigate misalignment issues in the 3RScan dataset, thereby improving the alignment between the ground truth bounding boxes and the predicted ones. This alignment enhancement is crucially important for accurate evaluation and reliable performance measurement in 3D visual grounding.

\begin{figure*}[hbtp]
 \centering
 \includegraphics[keepaspectratio, scale=0.32]
      {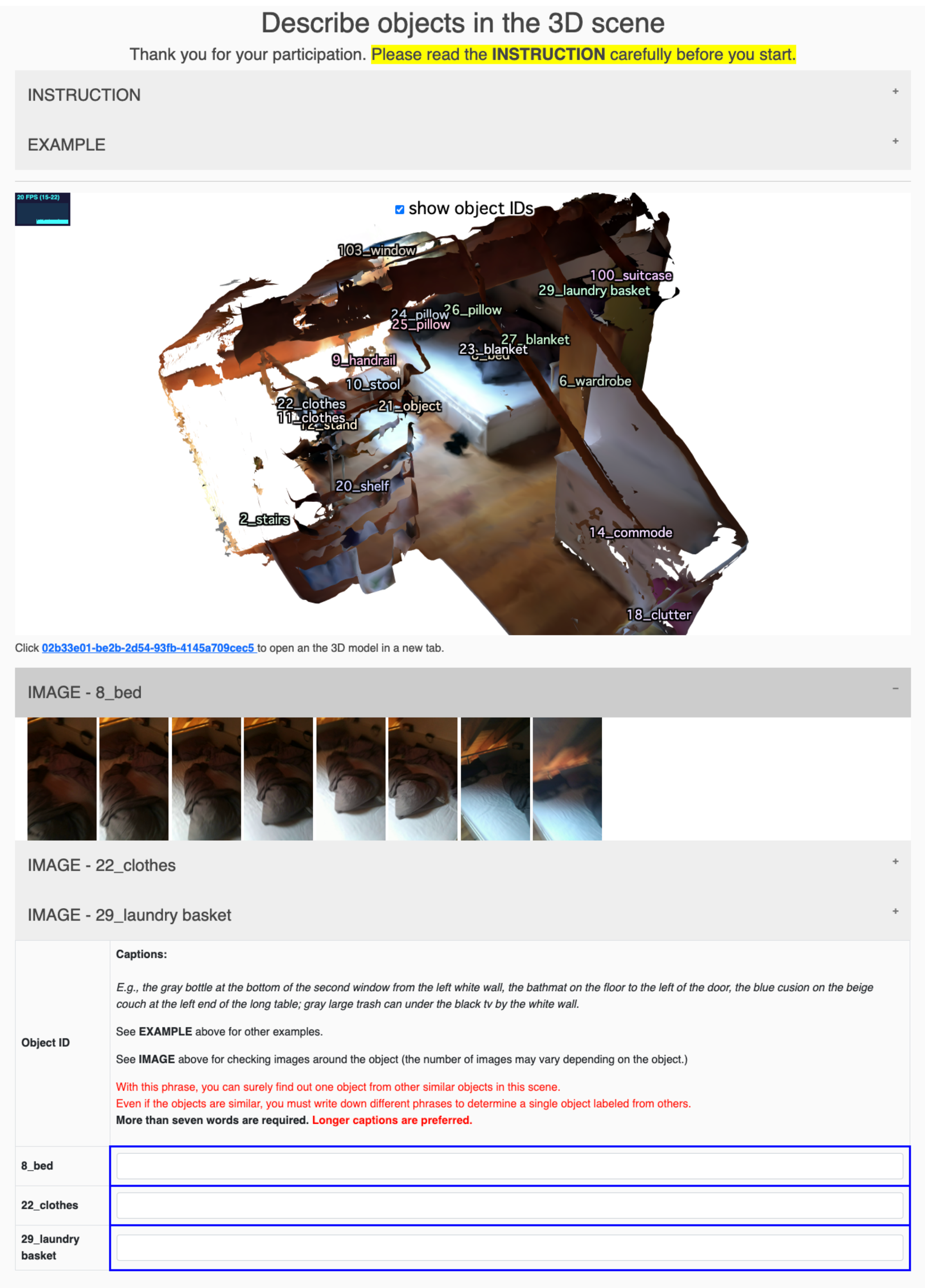}
 \caption{\small{Annotation website for RIORefer dataset creation. MTurk workers use this site to annotate the linguistic descriptions of objects in the given 3D scene.}}
 \label{fig:annotation_website}
\end{figure*}

%
%
\vspace{0.1cm}
\noindent
\textbf{Annotation Website.}
We developed an annotation website using the Amazon Mechanical Turk platform. This website was designed based on the approach described in the Supplemental Materials of the ScanRefer paper~\cite{chen2020scanrefer}.
Figure~\ref{fig:annotation_website} provides a screenshot of the annotation website, which portrays the user interface used by the workers.
After workers are presented with an object ID and object name, they must provide corresponding descriptions by observing a 3D model and object-related camera images displayed in the browser. This interactive visualization helps the workers describe the objects in the 3D scenes accurately.
 
\begin{figure*}[hbtp]
 \centering
 \includegraphics[keepaspectratio, scale=0.22]
      {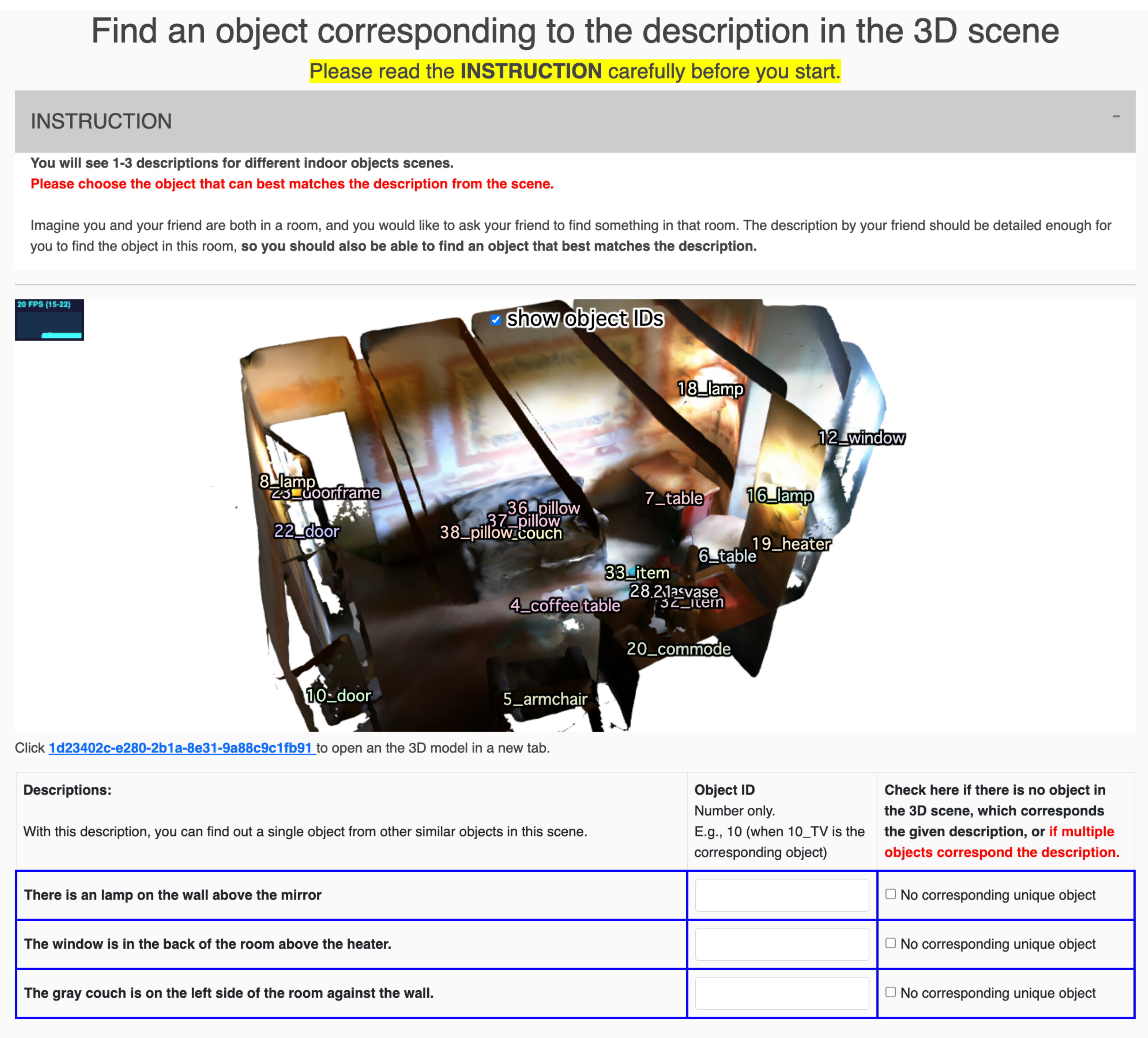}
\caption{\small{Annotation check website. MTurk workers use this site to validate the annotated descriptions while watching the given 3D model.}}
 \label{fig:check_website}
\end{figure*}

To ensure the quality of the annotations, we also developed the annotation check website for performing manual 3D visual grounding.
Figure~\ref{fig:check_website} shows a screenshot of the annotation check website.
Using this website, the workers verify the accuracy of the annotations by matching the descriptions with the corresponding object IDs.
If no object in the 3D scene matches the description alternatively if multiple objects match the description, then the workers are instructed to check the respective box.
Such incomplete or inaccurate descriptions are discarded. The corresponding objects are re-annotated using the annotation website shown in Figure~\ref{fig:annotation_website}.
By incorporating these annotation and annotation check websites into the data collection process, we ensure the quality and accuracy of the descriptions of the RIORefer dataset provided by the workers.

\vspace{0.1cm}
\noindent
\textbf{Human Performance.}
To evaluate the quality of the RIORefer dataset, we conducted a human performance assessment in 3D visual grounding using the annotation check website presented in Figure~\ref{fig:check_website}.

First, we randomly sampled 1,000 descriptions from the RIORefer dataset and assessed the percentage of descriptions that required refinement (designated as the {\em need refine rate}). 
Next, we removed the descriptions that needed refinement and selected an additional 1,000 descriptions from the refined subset. We then calculated the accuracy of manual 3D visual grounding ("grounding accuracy") for these 1,000 descriptions.
Results obtained for the RIORefer dataset showed a {\em need refine rate} of 0.05, indicating that only a small percentage of descriptions required refinement.
In addition, the {\em grounding accuracy} was measured as 0.903, indicating a high level of accuracy in manual 3D visual grounding.
For comparison, we also assessed the {\em need refine rate} and {\em grounding accuracy} of the ScanRefer dataset. The {\em need refine rate} for ScanRefer was 0.181, indicating a higher percentage of descriptions requiring refinement than RIORefer. The {\em grounding accuracy} of ScanRefer was measured as 0.797, indicating a slightly lower accuracy than that found for RIORefer.
These results demonstrate that the RIORefer dataset is a high-quality 3D visual grounding dataset that provides reliable and accurate annotations for training and evaluating 3D visual grounding methods.

\subsection{Dataset Analysis} 
\begin{figure*}[hbtp]
 \centering
 \includegraphics[keepaspectratio, scale=0.5]
      {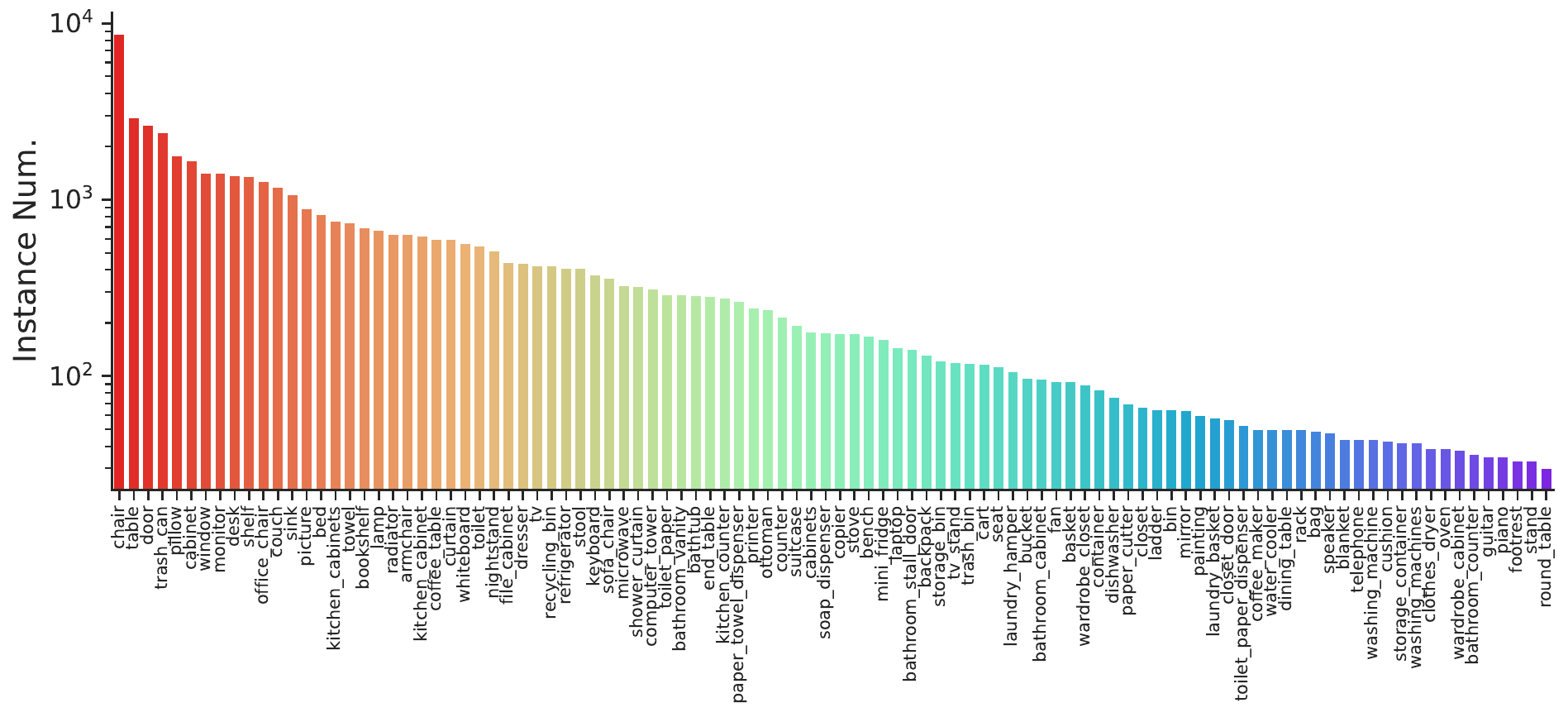}
 \caption{\small{Class category distribution for ScanRefer showing the number of instances per category. We use the most frequent 100 categories ordered by the number of instances per category on ScanRefer.}}
 \label{fig:scanrefer_inst_num}
\end{figure*}

\begin{figure*}[hbtp]
 \centering
 \includegraphics[keepaspectratio, scale=0.5]
      {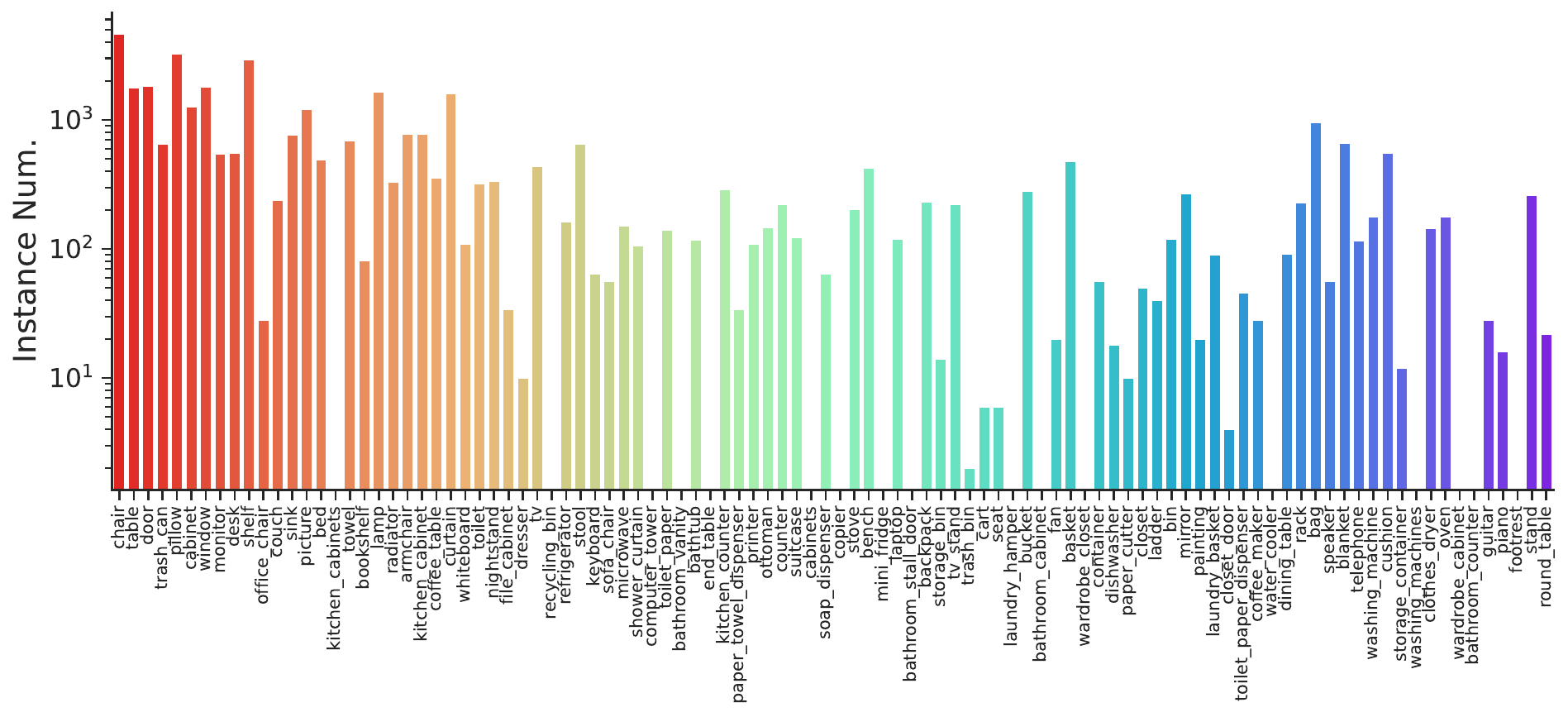}
 \caption{\small{Class category distribution for RIORefer showing the number of instances per category. We use the most frequent 100 categories ordered by the number of instances per category on ScanRefer.}}
 \label{fig:riorefer_inst_num}
\end{figure*}

We analyzed and compared the class category distribution of the ScanRefer and RIORefer datasets to emphasize their differences.
The distribution is presented in Figure~\ref{fig:scanrefer_inst_num} for ScanRefer and in Figure~\ref{fig:riorefer_inst_num} for RIORefer.
It is noteworthy that we used the most frequent 100 categories ordered by the number of instances per category on ScanRefer.
Comparison of the two figures reveals that ScanRefer and RIORefer exhibit distinct distributions related to object appearances.
Some objects (e.g., `computer tower,' `mini fridge,' and `laundry hamper') appear in ScanRefer, but they are absent in RIORefer.
This discrepancy brought the challenge of cross-dataset 3D visual grounding because it requires the ability to locate objects which might be unknown or occur infrequently in the training data.

\section{Additional Experiment Details}~\label{sec:supp_exp_setup}
We conducted additional experiments using the proposed baseline method and using various other methods
for further exploration of the performance of cross-dataset 3D visual grounding.

\subsection{{Additional Ablation Study}\label{sec:supp_ablation}}
\noindent
\textbf{Effects of joint learning of 2D and 3D.}
The proposed method (\textit{Ours} w/ joint\_train) uses 2D and 3D features and learned visual grounding with joint training using the loss $\mathcal{L}_{cls}(o_{2d})$ + $\mathcal{L}_{cls}(o_{3d})$ + $\mathcal{L}_{cls}([o_{2d} + o_{3d}])$.
Table~\ref{table:result_add_ablation} presents the performance of \textit{Ours} wo/ joint\_train using only $\mathcal{L}_{cls}([o_{2d} + o_{3d}])$.
The results exhibit a slight drop in performance compared to the joint training approach \textit{Ours} w/ joint\_train.
This finding indicates that predicting the target object with each 2D and 3D feature and their summation is more effective than with only the sum of 2D and 3D features.
Results demonstrated that the joint training strategy contributes to improvement of the performance of the visual grounding task in the cross-dataset setting.

\vspace{0.1cm}
\noindent
\textbf{Effects of many object categories.}
Instead of the 18 ScanNet benchmark classes used in ScanRefer~\cite{chen2020scanrefer}, we used ScanNet200 classes~\cite{rozenberszki2022language} for object detection labels to ascertain how the number of object categories affects the Cross3DVG performance.
These 200 object categories of ScanNet200 are divided into sets of 66 (\textit{head}), 68 (\textit{common}), and 66 (\textit{tail}) categories based on their respective frequencies of the numbers of labeled surface points.
Table~\ref{table:result_head_common_tail} shows
the results obtained using different sets of object categories (head, head+common, head+common+tail) for the object detection labels show some variations in performance over `multiple' and `unique' subsets of the Cross3DVG dataset.
\textit{Ours} ({head}+{common}+{tail}) outperformed \textit{Ours} ({head}) on the `multiple' subset, indicating that a larger set of object categories can improve performance in scenarios for which multiple objects must be grounded.
In contrast, \textit{Ours} ({head}) performed well over \textit{Ours} ({head}+{common}+{tail}) on the `unique' subset, suggesting that a smaller set of object categories can be effective for scenarios where grounding unique objects is necessary.
However, the performance of other methods did not consistently follow this trend across the subsets, indicating that the number of object categories on the Cross3DVG performance might vary depending on the specific method and on the dataset's characteristics.

%
%
\setlength{\tabcolsep}{2pt}
\begin{table}[t]
\begin{center}
	\footnotesize
	\begin{tabular}{lcc|cc}
        \toprule
    & \multicolumn{2}{c}{\scriptsize ScanRefer$\rightarrow$RIORefer} & \multicolumn{2}{c}{\scriptsize RIORefer$\rightarrow$ScanRefer} \\
    \midrule
Model  & {\scriptsize Acc@.25}  & {\scriptsize Acc@.5} & {\scriptsize Acc@.25}  & {\scriptsize Acc@.5} \\
\midrule
Ours (w/o joint\_train) & 24.25 & 12.64 & 32.58 & 19.05 \\
Ours (w/ joint\_train) & 24.78 & 13.34 & 33.34 & 19.48 \\
\bottomrule
	\end{tabular}
	\vspace{-0.2cm}
    \caption{\small{Ablation study of the proposed baseline method with and without the 2D--3D joint training.}}
    \label{table:result_add_ablation}
\end{center}
\vspace{-0.4cm}
\end{table}

%
%
\setlength{\tabcolsep}{2pt}
\begin{table}[t]
\begin{center}
	\footnotesize
	\begin{tabular}{lcc|cc}
        \toprule
    & \multicolumn{2}{c}{\scriptsize ScanRefer$\rightarrow$RIORefer} & \multicolumn{2}{c}{\scriptsize RIORefer$\rightarrow$ScanRefer} \\
    \midrule
Model & {\scriptsize Acc@.25}  & {\scriptsize Acc@.5} & {\scriptsize Acc@.25}  & {\scriptsize Acc@.5} \\
\midrule
VoteNet+MLP (H)  & 19.84 & 10.07 & 26.38 & 15.53 \\
VoteNet+MLP (H+C)  & 19.84 & 10.06 & 27.36 & 15.67 \\
VoteNet+MLP (H+C+T) & 19.69 & 9.91 & 27.27 & 15.99 \\
\midrule
DETR3D+Trans (H) & 23.49 & 12.58 & 31.42 & 18.97 \\
DETR3D+Trans (H+C) & 23.57 & 12.48 & 31.88 & 19.01 \\
DETR3D+Trans (H+C+T) & 23.27 & 12.31 & 31.33 & 18.60 \\
\midrule
Ours (H)  & 24.44 & 13.15 & 33.18 & 20.45 \\
Ours (H+C) & 24.01 & 12.70 & 33.58 & 20.02 \\
Ours (H+C+T) & 24.78 & 13.34 & 33.34 & 19.48 \\
\bottomrule
	\end{tabular}
    \caption{Ablation study of the proposed baseline method using different object category sets. Here, H, C, and T respectively denote the categories of head, common, and tails.}
    \label{table:result_head_common_tail}
\end{center}
\vspace{-0.4cm}
\end{table}

\subsection{Additional qualitative results}~\label{sec:supp_qualitative}
\begin{figure*}[hbtp]
 \centering
 \includegraphics[keepaspectratio, scale=0.9]
      {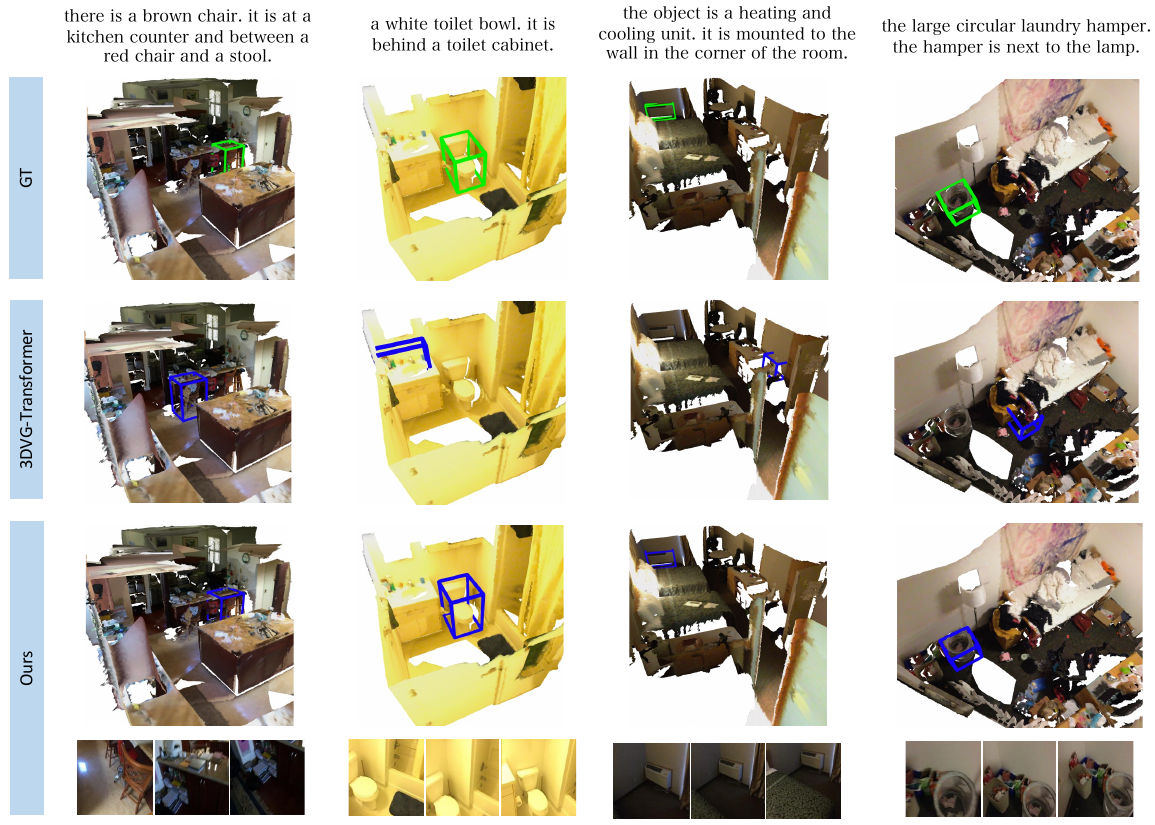}
 \caption{\small{Qualitative results from \textit{3DVG-Transformer} and \textit{Ours}. The GT (ground truth) boxes are marked in green. The predicted boxes having an IoU score higher than 0.5 are marked in blue. The text at the top of the image describes 3D visual grounding. Images at the bottom of \textit{Ours} were retrieved based on our method's coordinates of object proposals.}}
\label{fig:supp_qualitative}
\end{figure*}

In Figure~\ref{fig:supp_qualitative}, we present qualitative results for comparison of the \textit{3DVG-Transformer} and our proposed baseline method. These results specifically feature the models trained using RIORefer and subsequently tested with ScanRefer.
The results demonstrate that our method correctly predicts the target object according to the descriptions, which \textit{3DVG-Transformer} fails to predict. 
In the first column, for example, whereas \textit{3DVG-Transformer} predicts a different chair, our method identifies the correct chair successfully and then retrieves relevant multi-view images. These capabilities highlight the discriminative ability of our method in cross-dataset visual grounding using multi-view CLIP features.
Furthermore, our proposed method can predict objects that are not present in the training dataset.
As shown in the last column, \textit{3DVG-Transformer} fails to predict the correct object (`laundry hamper'), which is not included in the RIORefer training data but which appears in the ScanRefer dataset.
Our method used prior knowledge captured by CLIP and identified the target object using the retrieved multi-view images.
Overall, these qualitative results demonstrate the effectiveness of our proposed method for accurate prediction of target objects and for using multi-view images and CLIP's prior knowledge to accomplish cross-dataset visual grounding tasks.

\end{document}